\pdfoutput=1

\documentclass[11pt]{article}

\usepackage[]{acl}

\usepackage{times}
\usepackage{latexsym}

\usepackage[T1]{fontenc}

\usepackage[utf8]{inputenc}

\usepackage{microtype}

\usepackage{inconsolata}

\usepackage{babel} 
\usepackage{graphicx}
\usepackage{float} 
\usepackage[subrefformat=parens]{subcaption}
\usepackage{amsmath}

\usepackage{multirow}
\usepackage{tabularx}
\usepackage{makecell}
\usepackage{booktabs}

\usepackage{hyperref}
\usepackage[nameinlink]{cleveref}

\newcolumntype{K}[1]{>{\centering\arraybackslash}p{#1}} 

\title{ 
A multi-level multi-label text classification dataset of 19\textsuperscript{th} century Ottoman and Russian literary and critical texts  }

\author{Gokcen Gokceoglu$^1$, Devrim Cavusoglu$^1$, Emre Akbas$^1$, Ozen Nergis Dolcerocca$^2$ \\
$^1$Department of Computer Engineering, Middle East Technical University \\ 
$^2$Department of Modern Languages, Literatures, and Cultures, University of Bologna}

\begin{document}
\maketitle
\begin{abstract}
This paper introduces a multi-level, multi-label text classification dataset comprising over 3000 documents. The dataset features literary and critical texts from 19th-century Ottoman Turkish and Russian. It is the first study to apply large language models (LLMs) to this dataset, sourced from prominent literary periodicals of the era. The texts have been meticulously organized and labeled. This was done according to a taxonomic framework that takes into account both their structural and semantic attributes.  Articles are categorized and tagged with bibliometric metadata by human experts. We present baseline classification results using a classical bag-of-words (BoW) naive Bayes model and three modern LLMs: multilingual BERT, Falcon, and Llama-v2. We found that in certain cases, Bag of Words (BoW) outperforms Large Language Models (LLMs), emphasizing the need for additional research, especially in low-resource language settings. This dataset is expected to be a valuable resource for researchers in natural language processing and machine learning, especially for historical and low-resource languages. The dataset is publicly available\footnote{\href{https://huggingface.co/nonwestlit}{https://huggingface.co/nonwestlit}. The dataset, code, and trained models are released under Apache 2.0 License.}.
\end{abstract}

\section{Introduction}
\label{sec:intro}

\begin{figure*}
\begin{subfigure}[b]{.48\linewidth}
\centering
\includegraphics[width=\linewidth]{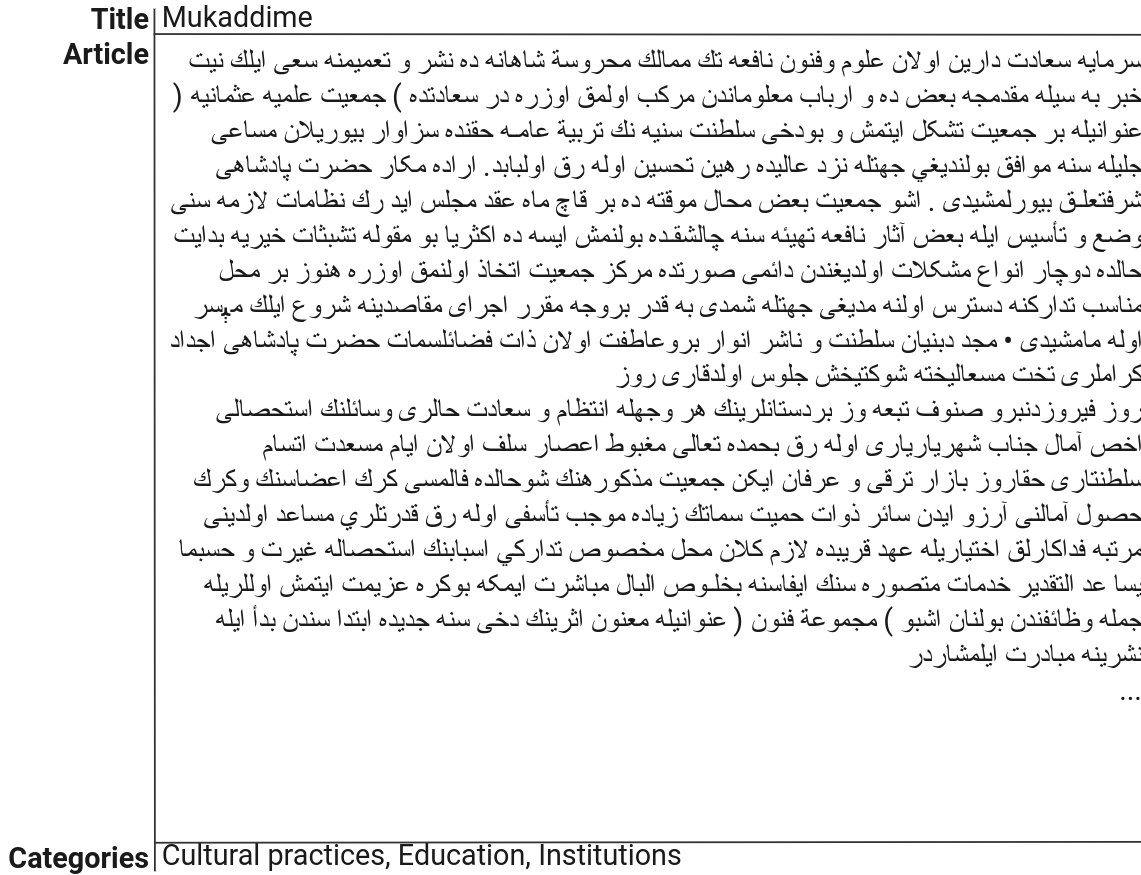}
\caption{An Ottoman example article.}\label{sfig:ottoman-example}
\end{subfigure}\hfill
\begin{subfigure}[b]{.48\linewidth}
\centering
\includegraphics[width=\linewidth]{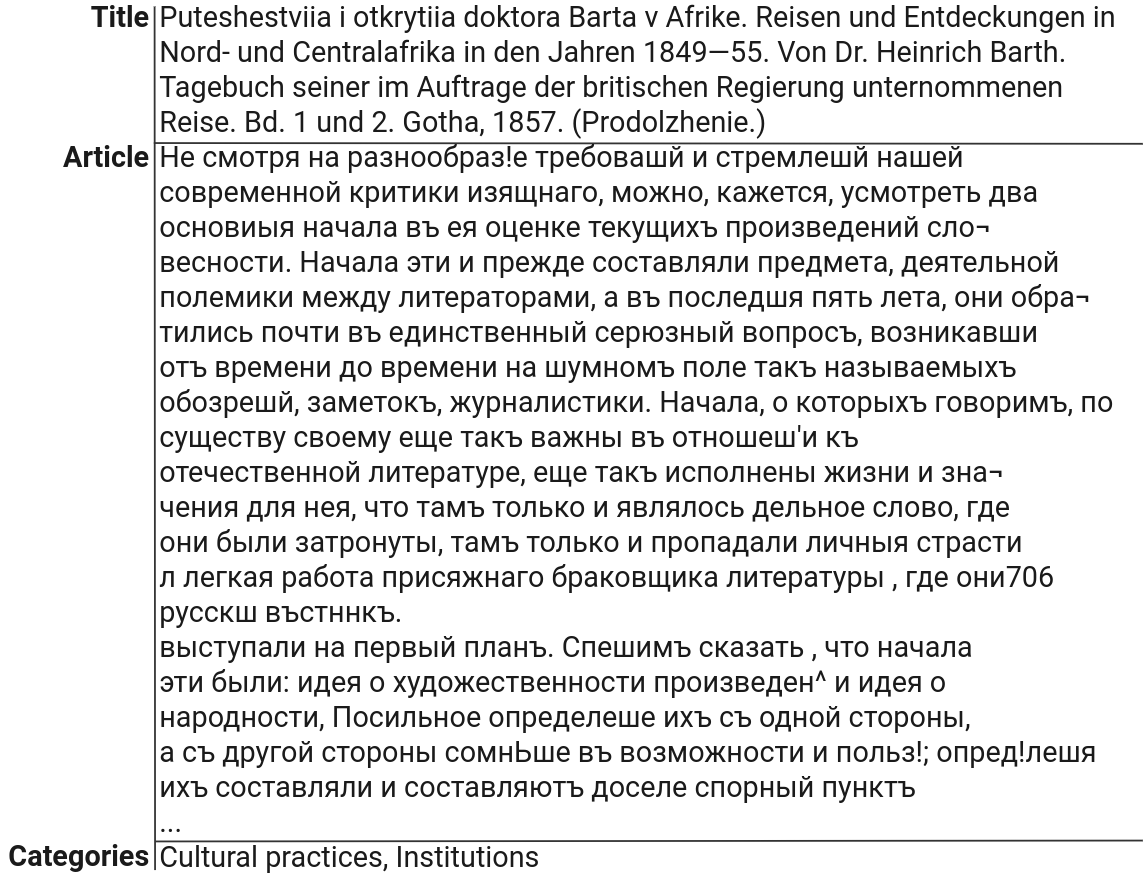}
\caption{A Russian example article.}\label{sfig:russian-example}
\end{subfigure}
\caption{A training instance from Ottoman \textbf{(a)} and Russian \textbf{(b)} collection samples from third level (i.e. Cultural Discourse $\rightarrow$ Modernization Subject) . The articles are truncated for better visual appearance.}
\label{fig:examples}
\end{figure*}

In recent years, there has been remarkable progress in natural language processing with the introduction of large language models \cite{vaswani2017attention,radford2018improving, devlin2018bert}, and well-curated, massive pre-training datasets. However, this progress has disproportionately benefited high-resource languages. For low-resource languages, the effectiveness of large language models is compromised by a number of issues including data scarcity that hinders generalization performance; tokenization processes that result in an inadequate representation of word meaning, and the bias in digital texts toward particular topics. To address these issues, there has been an increased focus on developing more inclusive and comprehensive models, introducing diverse datasets, and establishing benchmarks.

Low-resource languages, often comprising ancient or historical linguistic forms, are a key area of inquiry in computational linguistics characterized by linguistic diversity and evolutionary trajectories over time. However, the archival preservation of written documents poses a notable challenge to computational efforts, particularly in terms of resource availability. This challenge is even present in the case of relatively recent historical languages from the 19\textsuperscript{th} century, where not only limited textual resources but also non-standardized writing systems hinder computational analysis and linguistic research efforts. 


With the recent advancements in deep learning research, the effort to digitize and apply machine comprehension on these type of archival resources has become a growing field of inquiry, fostering inter-disciplinary collaborations between humanities, social sciences, and computer science. One example of such endeavor that aims to overcome the limitation of lack of resources, is a large-scale project focusing on aggregation, refinement, and digitizing historical newspapers for The European Library\footnote{\hyperlink{https://www.theeuropeanlibrary.org/}{https://www.theeuropeanlibrary.org/}} and Europeana\footnote{\hyperlink{http://www.europeana.eu/}{http://www.europeana.eu/}} funded by the European Commission \cite{pekarek2012europeana}. In the project, several refinement methods were used on the raw content including, but not limited to Optical Character Recognition (OCR), Optical Layout Recognition, and Named Entity Recognition. The project included many languages and contributions from many national libraries having over 16 million items. It largely covered main Western European languages. There were, therefore,  several limitations for those languages that were out of the scope of the project, which resulted in limitations for refinement techniques for those particular non-European languages (e.g. Ottoman) \cite{europeana2013digitised}.

In order to address the limitations of digitization and refinement of non-European languages, this paper focuses on historical non-Western languages, primarily Ottoman and Russian literature texts from the 1830s to the 1910s, the most prolific years for periodical publications in the long 19\textsuperscript{th} century. The rise of mass readership in periodical press roughly corresponds to these dates across all cultures with mass printing presses and (proto) print capitalism \cite{vincent2000rise}. While Russian is not considered a low-resource language, computational efforts, particularly for nineteenth-century texts, remain limited \cite{martinsen1997literary}. Ottoman Turkish, on the other hand, presents a unique case. It was a written language distinct from the vernacular, featuring vocabulary and syntactic elements borrowed from Persian and Arabic \cite{kabacali2000bacslangicindan}. The orthography was also not standardized, being written in the Arabo-Persian alphabet with variations in spelling and punctuation. This is in a stark contrast to modern Turkish which has undergone systematic phonocentric standardization toward the vernacular and is written in the Latin alphabet. Therefore, any computational study conducted in modern Turkish would have significant limitations in Ottoman Turkish, as in any non-standardized language \cite{ryskina2022learning}. 


In this paper, we provide an open-access, multi-label text classification dataset consisting of Ottoman and Russian literary texts from the 19\textsuperscript{th} century. The dataset collection was prepared with due diligence from data collection to labeling. We used an OCR pipeline and developed a web-based platform for a team of experts to label the digitized content. Our dataset collection has a specific hierarchical structure (see \Cref{apx:structure}) and is annotated according to pre-determined categories as either single-label or multi-label. Example instances from both the Ottoman and Russian datasets are given in \Cref{fig:examples}.

We expect that these text classification datasets will prove instrumental in advancing future research endeavors. The availability of these datasets will enable researchers to train a wide array of models tailored to specific applications. This includes the automated categorization of digitized resources, offering enhanced efficiency and accuracy in managing digital information. We also note that the Russian and Ottoman datasets consist mainly of literary and critical texts, which offer a potential for unsupervised learning.



\section{Related Work}
\label{sec:related-work}
A key milestone behind the recent breakthroughs in NLP is the extensive pre-training datasets. These datasets are collected from digital sources, where Western languages, predominantly English, are the primary languages represented. Specifically, by January 2024, W3Techs estimates that Western languages make up $83.1\%$ and English $51.7\%$, of the websites \cite{w3tech2024lang}. 

In recent years, efforts have been made to bridge this gap by introducing of multilingual and low-resource datasets and developing methods to improve model performance. These studies have targeted different NLP tasks, including machine translation \cite{guzman2019flores, mueller2020analysis}, text classification \cite{cruz2020establishing, zaikis2023pre, fesseha2021text, liebeskind2020deep}, and part of speech tagging \cite{csahin2019data}. Efforts have been made to develop new augmentation techniques \cite{csahin2019data} and training strategies \cite{tang2018improving}. 

\citet{gopidi-alam-2019-computational} proposed an English dataset spanning the late 19\textsuperscript{th} to late 20\textsuperscript{th} centuries to examine the shift between poetry and prose resources. \citet{cruz2020establishing} introduced two datasets containing over 14,000 samples in Filipino for binary and multi-label classification tasks.  
\citet{guzman2019flores} proposed datasets for Nepali-English and Sinhala-English translations, which provide benchmarks for evaluating methods trained on low-resource language pairs. \citet{mueller2020analysis} used the Bible for machine translation tasks in 1107 languages, creating a multilingual corpus by varying the number and relatedness of source languages. 
\citet{fesseha2021text} provided a dataset curated for multi-label classification and a pre-training dataset in Tigrinya, a Semitic language spoken primarily in Eritrea and northern Ethiopia. 
\citet{regatte2020dataset} introduced a dataset for sentiment analysis in Telagu, which is one of the most spoken languages in India, but lacks adequate digital resources.
\citet{bansal2021low} presented a dataset in Sumerian, one of the earliest known written languages, and proposed a cross-lingual information extraction pipeline.
As a low-resource language that is no longer in use, Ottoman Turkish has been minimally used in NLP research. One such study used traditional machine learning methods to classify poems from the 15\textsuperscript{th} to 19\textsuperscript{th} centuries by period and genre \cite{can2012automatic, can2013automatic}. Besides these low-resource dataset proposals and models trained on top of them, there are studies on improving training/fine-tuning approaches for low-resource languages in a multilingual setting \cite{adapt2023lankford, ogueji-etal-2021-small}.

In this paper, we present the first expert-curated and labeled dataset in Ottoman Turkish and Russian and propose the first study employing LLMs for multi-level classification tasks in both languages.

\section{Data Collection}
\label{subsec:data-collection}

Our dataset collection procedure had two steps:  article curation and labeling by human experts.

\subsection{Article Curation}

Our main resources are hard copies and soft copies of texts from literary periodicals from the 1830s to 1910s, gathered from main archival resources including national and imperial libraries and digitized corpus. Experts determined a range of significant literary journals from the period, paying special attention to the wide representation of the ideological and cultural spectrum of the period. The hard or digitized copies of the selected issues of these journals were then processed and segmented into sections for each entry in a particular issue. 

Due to the predominant presence of hard copy resources, particularly in Ottoman Turkish, it was necessary to digitize some of these documents to make them processable by computers. To this end, we have developed a rigorous digitization protocol aimed at transforming physical articles into digital content. This process is structured into two sequential stages: OCR followed by correction and refinement of the digitized text to ensure fidelity to the original content. OCR process for documents in Ottoman Turkish is conducted through Google Lens\footnote{\href{https://lens.google/}{https://lens.google/}}, as the other OCR programs we experimented with, Abbyy \cite{abbyy_ocr_sdk} and Tesseract (Arabic / Persian) \cite{kay2007tesseract}, yielded low accuracy. Each PDF document page is converted to images, uploaded to Google Photos automatically via Google Cloud, and digitized by archive workers. The polishing and corrections on the digitized documents were conducted and completed by the team of experts also designated for the labeling task. We chose not to perform a formal evaluation due to time constraints and resource limitations, but the measures implemented have provided us with a high degree of confidence in the transcription accuracy. 

We built a system with a web-based user interface where users can interact and label the digitized documents accordingly. The designated team of experts was signed up to the system, where each user can interact with a document reader and is able to view the original images of the same document that are fed into the OCR pipeline. This was done to reduce errors in the correction phase of the digitized text.  

\subsection{Labeling}
\label{ssec:labeling}

As reading, understanding and actively interpreting these texts are required for their annotation, we built a team of experts for each area, based on their academic background, linguistic skills, and scientific output regarding the nineteenth-century literary culture. The team of experts is assigned tasks to segment and label these digitized articles to form a text classification dataset. The annotations underwent a randomized cross-check process by the designated expert-leader of each language. The experts also tagged documents with bibliometric metadata in addition to labeling the determined category of that article type. This is crucial for the digitization process such that, the dataset and the database we built in parallel could be searchable in a more comprehensive way with bibliometric information which we believe will ease the search process for researchers. 

The taxonomy represents the formal and thematic elements in the cultural and aesthetic world of the period. Texts are first categorized based on their form, i.e. poetry, critical article, or short story. Critical articles, labeled ``cultural discourse,'' are then categorized according to the subject of their content.  

\section{Dataset Analysis}
\label{subsec:dataset-analysis}

\begin{table}
\small
\centering
    \begin{tabular}{ l r r }
        \toprule
        \textbf{Category} & \textbf{Ottoman} & \textbf{Russian} \\
        \midrule
        \textbf{Literary Text} & 481 & 357\\
        |--- Short Story & 99 & 44\\
        |--- Poetry & 278 & 244\\
        |--- Translated Text & 4 & 3\\
        |--- Novel & 57 & 33\\
        |\_\_ Play & 8 & 15\\
        \midrule
        \textbf{Cultural Discourse} & 929 & 475\\
        \textbf{\textit{Types}} \\
        |--- Article & 796 & 272\\
        |--- Review & 91 & 142\\
        |--- Biography & 74 & 37\\
        |--- Letter & 77 & 65\\
        |--- Manifesto & 30 & 7\\
        |--- Travel writing & 11 & 14\\
        \textbf{\textit{Subjects}} \\
        |--- \textbf{Literature} & 489 & 173\\
        |--- \textbf{Philosophy} & 574 & 236\\
        |--- Politics & 105 & 7\\
        |--- Translation & 53 & 16\\
        |--- \textbf{Modernization} & 152 & 132\\
        |--- \textbf{Identity} & 136 & 32\\
        |\_\_ \textbf{Language} & 169 & 28 \\
        \midrule
        \textbf{Other} & 409 & 226\\
        \midrule
        \textbf{Total} & 1819 & 1058\\
        \bottomrule
    \end{tabular}
    \caption{Taxonomy and tree structure for the Ottoman and Russian dataset for first and second level categories. Subcategories having lower-level categories are marked in bold.}
    \label{tab:taxonomy1-2}
\end{table}

Our dataset comprises articles from Russian and Ottoman sources spanning the late 19\textsuperscript{th} and early 20\textsuperscript{th} centuries. The structures of datasets have a hierarchical scheme, where the datasets follow 4 levels of categorization. The articles are primarily grouped into 3 categories based on their relationship to literary and cultural phenomena, namely ``Literary Text,'' ``Cultural Discourse'' and ``Other,'' which marks the first level of categorization (L1). ``Other'' refers to everything that remains outside of literary or cultural texts, such as news articles, advertisements, and obituaries. Literary Text and Cultural Discourse categories also have lower-level categories. There is no lower-level category for the articles in the class ``Other.'' The article counts and the names of the subcategories are shown in \Cref{tab:taxonomy1-2}.

The categorization goes up to the fourth level of categories, and the tree structure displaying these lower levels is given in \Cref{tab:taxonomy3-4}. 

Within the Ottoman dataset, we have a collection of 1,819 articles from 685 journals, with an average of 1,005.29 words, and 43.4 sentences per article. The Russian dataset contains 1,058 articles from 198 journals, with an average of 4,630.69 words and 212.26 sentences per article.

\begin{table}
\small
\centering
    \begin{tabular}{ p{4cm} r r }
        \toprule
        \textbf{Category} & \textbf{Ottoman} & \textbf{Russian} \\
        \midrule
        \textbf{Literature Genre} & 490 & 173\\
        |--- Poetry & 152 & 55\\
        |--- Theater & 37 & 48\\
        |--- Prose Fiction & 86 & 36\\
        \textbf{Literary Movement} & 490 & 173\\
        |--- Decadence & 20 & 1\\
        |--- Symbolism & 28 & -\\
        |--- Romanticism & 70 & 21\\
        |--- Traditionalism & 33 & -\\
        |--- Realism & 49 & 20\\
        |--- Classicism & 36 & 16\\
        |--- Naturalism & 23 & 5\\
        |--- Modernism & 52 & -\\
        |\_\_ Sentimentalism & 2 & 8\\
        \midrule
        \textbf{Philosophy} & 576 & 236\\
        |--- Political Philosophy & 202 & 173\\
        \hspace{12px}|--- State & 57 & 70\\
        \hspace{12px}|--- Race & 25 & 4\\
        \hspace{12px}|--- Empire/Colony & 18 & 16\\
        \hspace{12px}|--- Nation/Society &  133 & 132\\
        \hspace{12px}|--- Economy & 46 & 63\\
        \hspace{12px}|--- Law & 32 & 52\\
        \hspace{12px}|\_\_ Class/Capital & 12 & 38\\
        |--- Epistemology/Ontology & 113 & -\\
        |--- Ethics & 174 & 35\\
        \hspace{12px}|--- Religion/Secularism & 75 & 20 \\
        \hspace{12px}|\_\_ Morality & 112 & 23\\
        |--- Aesthetics & 105 & 32 \\
        \hspace{12px}|--- Didacticism & 56 & 21 \\
        \hspace{12px}|\_\_ Aestheticism & 49 & 29\\
        |--- Philosophy of History & 152 & 39 \\

        |\_\_ Movement & 229 & 25\\
        \hspace{12px}|--- Enlightenment & 49 & 4\\
        \hspace{12px}|--- Nationalism & 120 & 7\\
        \hspace{12px}|--- Materialism & 31 & 6\\
        \hspace{12px}|--- Woman’s Question & 25 & 6\\
        \hspace{12px}|--- Idealism & 26 & 8\\
        \hspace{12px}|--- Orientalism & 15 & -\\
        \hspace{12px}|\_\_ Marxism & 4 & 6\\
        \midrule
        \textbf{Modernization} & 152 & 132\\
        |--- Cultural practices & 42 & 36\\
        |--- Education & 70 & 56\\
        |--- Print culture & 35 & 34\\
        |--- Institutions & 39 & 19\\
        |\_\_ Urbanization & 22 & 7\\
        \midrule
        \textbf{Identity} & 136 & 32\\
        |--- Localism & 102 & 28\\
        |\_\_ Westernizer & 65 & 22\\
        \midrule
        \textbf{Language} & 169 & 28\\
        |--- Language reforms & 62 & 1\\
        |--- Stylistics & 55 & 12\\
        |\_\_ Linguistics & 77 & 19\\
        \bottomrule
    \end{tabular}
    \caption{Taxonomy for the Ottoman and Russian dataset for third and fourth level categories.}
    \label{tab:taxonomy3-4}
\end{table}

\section{Modeling experiments}
\label{sec:modeling-experiments}

We conducted multiple experiments using Ottoman and Russian datasets to establish baseline results for classification tasks.

\subsection{Experimental Setup}
\label{subsec:exp-setup}

\begin{table*}
\small
\centering
    \begin{tabular}{c | K{1.25cm} | l | c c p{1.25cm} p{1.25cm} p{1.25cm}}
        \toprule
        \textbf{Dataset} & \textbf{Category} & \textbf{Models} &  \textbf{Acc.} & \textbf{mAP (macro)} & \textbf{mAP (weighted)} & \textbf{F1 (macro)} & \textbf{F1 (weighted)} \\
        \midrule
        Ottoman L1 & - & \makecell[c]{Llama-2-7b \\ Falcon-7b \\ mBERT \\ BoW + NB} & \makecell[c]{77.99 \\ 77.93\\ 78.91\\ \textbf{79.53}} & \makecell[c]{79.15\\ 77.93\\ 78.91\\ \textbf{79.53}} & \makecell[c]{\textbf{81.54}\\ 81.08\\ 81.00\\ 78.75} & \makecell[c]{74.54\\ \textbf{82.44}\\ 71.04\\ 64.75} & \makecell[c]{77.07\\ 62.65\\ \textbf{77.65}\\ 68.05}\\
        \midrule
        Russian L1 & - & \makecell[c]{Llama-2-7b \\ Falcon-7b \\ mBERT \\ BoW + NB} & \makecell[c]{\textbf{79.11} \\ 77.22\\ 78.92\\ 77.35} & \makecell[c]{\textbf{83.44}\\ 75.10\\ 80.53\\ 71.45} & \makecell[c]{\textbf{87.40}\\ 80.22\\ 84.30\\ 75.81} & \makecell[c]{75.26\\ 58.79\\ \textbf{78.56}\\ 70.67} & \makecell[c]{\textbf{79.89}\\ 71.79\\ 74.01\\ 74.85}\\
        \midrule
        \midrule
        Ottoman L2 & literary text type & \makecell[c]{Llama-2-7b \\ Falcon-7b \\ mBERT \\ BoW + NB} & \makecell[c]{53.61\\ 52.94\\ 55.89\\ \textbf{82.92}} & \makecell[c]{50.05\\ 50.23\\ 31.16\\\textbf{87.31}} & \makecell[c]{50.23\\ 50.23\\ 65.39\\\textbf{92.50}} & \makecell[c]{51.5\\ 50.41\\ 30.84\\ \textbf{78.27}} & \makecell[c]{53.04\\ 50.41\\ 52.78\\ \textbf{83.84}}\\
        \midrule
        Russian L2 & literary text type & \makecell[c]{Llama-2-7b \\ Falcon-7b \\ mBERT \\ BoW + NB} & \makecell[c]{68.50\\ 50.00\\ 7.41\\ \textbf{69.19}} & \makecell[c]{\textbf{54.52}\\ 37.60\\ 35.03\\ 38.95} & \makecell[c]{\textbf{88.25}\\ 76.96\\ 73.65\\ 85.95} & \makecell[c]{\textbf{39.13}\\ 24.43\\ 3.45\\ 24.44} & \makecell[c]{\textbf{71.25}\\ 53.95\\ 1.02\\ 65.69}\\
        \bottomrule
    \end{tabular}
    \caption{Model performances for first-level (L1) and second-level (L2) single-label classification.}
    \label{tab:result-single}
\end{table*}

\paragraph{Modeling Setup.}

The selection of appropriate language models is crucial for creating an effective experimental setup.
Factors such as multilingual support, extensive pre-training corpora,  and models including billion-parameter LLMs \cite{ouyang2022training, falcon40b, touvron2023llama} should be considered.  
Given these considerations, we chose two recent open-source LLMs with large and preferably multilingual pretraining data, namely Llama-2 \cite{touvron2023llama2} and Falcon \cite{falcon40b}. These models were chosen to maximize efficiency and exploit the knowledge they have from their large pre-training datasets. 
Llama-2 utilizes a massive dataset collected from publicly available resources, while Falcon uses the RefinedWeb Dataset 
\cite{refinedweb}, a cleaned text dataset from 
CommonCrawl
\footnote{\hyperlink{https://commoncrawl.org/}{https://commoncrawl.org/}}. Additionally, we included multilingual BERT (mBERT) \cite{devlin-etal-2019-bert} to explore the capabilities of a smaller-scale model in our experiments.

Due to the memory-intensive nature of the pre-trained large-scale transformer models, we opted for "Llama-2-7b"\footnote{\hyperlink{https://huggingface.co/meta-llama/Llama-2-7b-hf}{https://huggingface.co/meta-llama/Llama-2-7b-hf}} and "Falcon-7b"\footnote{\hyperlink{https://huggingface.co/tiiuae/falcon-7b}{https://huggingface.co/tiiuae/falcon-7b}} variants. For smaller scale model we chose "mBERT-base"\footnote{\hyperlink{https://huggingface.co/bert-base-multilingual-cased}{https://huggingface.co/bert-base-multilingual-cased}} variant for multilingual BERT. Llama-2-7b and Falcon-7b are the smallest variants available for these model families. We used Nvidia RTX 3090 and 4090 GPUs, both of which have 24 GB of vRAM. For efficient training on our compute resources, we opted for 4-bit quantization \cite{dettmers2022llmint8}. In addition, we did not train all model parameters, but instead used LoRa \cite{hu2022lora} and trained LoRa parameters only. For mBERT, we opted for linear probing and trained only the classification head, keeping the backbone frozen. We also set the floating point type as bfloat16.

In training, we used 2048 context window, a batch size of 4, and conducted a hyperparameter search on learning rate (ranging between $1.75\times 10^{-5}$ and $2\times 10^{-4}$) and weight decay (ranging between $10^{-4}$ and $10^{-2}$) values.

In addition to LLMs, we use a classical ``bag of word naive Bayes'' (BoW+NB) model to give a simpler baseline for models prior to LLMs. Each article in the dataset is represented by a binary vector, whose size corresponds to the number of unique words in the dataset of the corresponding classification task. For Russian, the dictionary size varies for different classification tasks, ranging from 163,615 to 242,305 words. For Ottoman, it spans from 55,507 to 179,873 words. To achieve classification tasks, we used a multinomial naive Bayes classifier.  

\paragraph{Fine-tuning over Chunks.}

Since articles can exceed 2048 tokens (our max. input length), we trained the model over chunks of articles for those that exceed this length, i.e., we applied chunking with non-overlapping sliding windows of length 2048.

\paragraph{Chunked Inference.}

At test/inference time, we applied chunking to the input article (at the same size as the trained model's max. input length, 2048), obtained the classification probabilities for each chunk, and then applied average pooling class-wise to obtain the final probabilities of the article.

\paragraph{Dataset.}

For all classification tasks, we used stratified sampling to split the dataset into training  (70\%), validation  (15\%), and test (15\%) sets. 
We eliminated the classes with less than 10 samples before splitting and did not use them in training.  
In addition, we did not conduct experiments on datasets where the sample size of the test is less than 15 instances.

\paragraph{Metrics.}

To evaluate the models, we reported accuracy (percentage of correct predictions), mean average precision (mAP), and F1 scores for single-label tasks. For multi-label tasks, we provided mAP, average precision (AP@0.5), average recall (AR@0.5), and average F1 (AF1@0.5) scores, all globally averaged at the 0.5 threshold. We utilized the scikit-learn package \cite{pedregosa2011sklearn} for both single-label and multi-label metric computations. For the mAP and F1 metrics, we reported both macro averages (unweighted mean across all classes, i.e., each class is treated equally) and weighted average scores (classes are weighted by their sample size), although we focused on weighted metrics. While we provide this set of metrics, we monitor the mAP (weighted) metric to assess model quality, as it is common to both single-label and multi-label tasks.

\subsection{Model Results}
\label{subsec:results-ottoman}

\begin{table*}
\small
\centering
    \begin{tabular}{c | K{2.25cm} | l | c c p{1.25cm} p{1.25cm} p{1.25cm}}
        \toprule
        \textbf{Dataset} & \textbf{Category} & \textbf{Models} & \textbf{mAP (macro)} & \textbf{mAP (weighted)} & \textbf{AP@0.5} & \textbf{AR@0.5} & \textbf{AF1@0.5} \\
        \midrule
        Ottoman L2 & cultural discourse subject & \makecell[c]{Llama-2-7b \\ Falcon-7b \\ mBERT \\ BoW + NB} & \makecell[c]{30.62 \\ 30.11\\ 30.74 \\ \textbf{42.44} } & \makecell[c]{52.02\\ 51.65\\ 53.14\\ \textbf{64.43} } & \makecell[c]{62.52\\ 65.74\\ 57.24\\ \textbf{72.43} } & \makecell[c]{58.77\\ 58.13\\ 60.08 \\ \textbf{70.21}} & \makecell[c]{60.59 \\ 61.70\\ 58.63\\ \textbf{71.30}}\\
        \midrule
        Ottoman L2 & cultural discourse type & \makecell[c]{Llama-2-7b \\ Falcon-7b \\ mBERT \\ BoW + NB} & \makecell[c]{18.88\\ 18.57\\ 19.70 \\\textbf{35.13} } & \makecell[c]{66.09\\ 63.73\\ 66.44 \\ \textbf{69.29}} & \makecell[c]{84.67\\ 84.76\\ 84.78\\ \textbf{87.23}} & \makecell[c]{70.03\\ 70.30\\ \textbf{70.48} \\70.08} & \makecell[c]{76.82\\ 76.82\\ 76.97\\ \textbf{77.72}}\\
        \midrule
        Russian L2 & cultural discourse subject & \makecell[c]{Llama-2-7b \\ Falcon-7b \\ mBERT \\ BoW + NB} & \makecell[c]{\textbf{46.94}\\ 26.35\\ 32.54 \\26.55 } & \makecell[c]{\textbf{69.60}\\ 45.06\\ 60.63\\ 36.04} & \makecell[c]{\textbf{69.13}\\ 42.65\\ 44.12\\ 41.34} & \makecell[c]{\textbf{57.14}\\ 29.59\\ 30.61 \\34.25} & \makecell[c]{\textbf{62.57}\\ 34.94\\ 36.14\\ 37.46}\\
        \midrule
        Russian L2 & cultural discourse type & \makecell[c]{Llama-2-7b \\ Falcon-7b \\ mBERT \\ BoW + NB} & \makecell[c]{\textbf{43.96}\\ 24.24\\ 32.51 \\36.32 } & \makecell[c]{\textbf{67.23}\\ 48.13\\ 59.51\\ 49.93} & \makecell[c]{48.15\\ 54.17\\ 54.17\\ \textbf{63.80}} & \makecell[c]{\textbf{46.43}\\ \textbf{46.43}\\ \textbf{46.43} \\ 45.89} & \makecell[c]{47.28\\ 50.00\\ 50.00\\ \textbf{53.38}}\\
        \midrule
        \midrule
        Ottoman L3 & philosophy subject & \makecell[c]{Llama-2-7b \\ Falcon-7b \\ mBERT \\ BoW + NB} & \makecell[c]{37.05\\ 36.28\\ 36.05 \\\textbf{55.20} } & \makecell[c]{40.61\\ 39.90\\ 39.54\\ \textbf{57.54}} & \makecell[c]{44.75\\ 43.49\\ 00.00\\ \textbf{68.14}} & \makecell[c]{30.68\\ 21.82 \\ 00.00 \\\textbf{37.55}} & \makecell[c]{36.40\\ 29.06\\ 00.00\\ \textbf{48.42}}\\
        \midrule
        Ottoman L3 & literary movement & \makecell[c]{Llama-2-7b \\ Falcon-7b \\ mBERT \\ BoW + NB} & \makecell[c]{33.02 \\ 25.96\\ 26.11 \\ \textbf{40.37} } & \makecell[c]{32.04 \\ 28.68\\ 29.14 \\ \textbf{43.09} } & \makecell[c]{ 36.95 \\ 42.85\\ 00.00 \\ \textbf{71.42}} & \makecell[c]{06.85\\ 01.09\\ 00.00 \\ \textbf{16.94} } & \makecell[c]{11.56 \\ 02.14\\ 00.00 \\ \textbf{27.39} }\\
        \midrule
        Ottoman L3 & literature subject genre & \makecell[c]{Llama-2-7b \\ Falcon-7b \\ mBERT \\ BoW + NB} & \makecell[c]{\textbf{59.18}\\ 50.85\\ 48.13 \\55.38 } & \makecell[c]{65.43\\ 56.47\\ 55.30 \\\textbf{69.13} } & \makecell[c]{65.60\\ 68.31\\ 70.59 \\ \textbf{80.00} } & \makecell[c]{53.36\\ 55.19\\ 58.54 \\ \textbf{73.33} } & \makecell[c]{58.85\\ 58.58\\ 64.00 \\ \textbf{76.52} }\\

        \midrule
        Ottoman L3 & modernization subject & \makecell[c]{Llama-2-7b \\ Falcon-7b \\ mBERT \\ BoW + NB} & \makecell[c]{39.13\\ 34.67\\ 27.12 \\\textbf{54.45} } & \makecell[c]{42.83\\ 37.86 \\ 38.42 \\\textbf{58.68} } & \makecell[c]{49.18\\ 47.82 \\ 00.00 \\\textbf{51.35} } & \makecell[c]{28.39\\ 12.50\\ 00.00 \\\textbf{36.53} } & \makecell[c]{34.48\\ 00.00\\ 00.00 \\\textbf{42.69} }\\

        \midrule
        Ottoman L3 & identity type & \makecell[c]{Llama-2-7b \\ Falcon-7b \\ mBERT \\ BoW + NB} & \makecell[c]{65.42\\ \textbf{70.45}\\ 56.12 \\65.94 } & \makecell[c]{66.48\\ 70.84\\ 57.64 \\\textbf{76.48} } & \makecell[c]{64.22\\ 67.28\\ \textbf{68.19} \\53.33 } & \makecell[c]{77.22\\ \textbf{91.22}\\ 60.00 \\55.81} & \makecell[c]{68.42\\ \textbf{74.55}\\ 63.83 \\ 54.54 }\\

        \midrule
        Russian L3 & philosophy subject & \makecell[c]{Llama-2-7b \\ Falcon-7b \\ mBERT \\ BoW + NB} & \makecell[c]{\textbf{61.68}\\ 30.73\\ 47.11 \\49.64 } & \makecell[c]{\textbf{69.94}\\ 49.42\\ 65.29 \\62.62 } & \makecell[c]{78.38\\ \textbf{74.29}\\ \textbf{74.29} \\60.78 } & \makecell[c]{\textbf{61.70}\\ 55.32\\ 55.32 \\42.95 } & \makecell[c]{\textbf{69.05}\\ 63.41\\ 63.41 \\49.99 }\\
        \midrule
        Russian L3 & literature subject genre  & \makecell[c]{Llama-2-7b \\ Falcon-7b \\ mBERT \\ BoW + NB} & \makecell[c]{40.08\\ 38.40\\ \textbf{53.55} \\52.13 } & \makecell[c]{40.07\\ 38.77\\ 54.15 \\\textbf{56.51} } & \makecell[c]{30.00\\ 28.57\\ 00.00 \\ \textbf{65.62} } & \makecell[c]{13.64\\ 9.09\\ 00.00 \\\textbf{70.00} } & \makecell[c]{18.75\\ 13.80\\ 00.00 \\\textbf{67.74} }\\
        \midrule
        Russian L3 & modernization subject & \makecell[c]{Llama-2-7b \\ Falcon-7b \\ mBERT \\ BoW + NB} & \makecell[c]{\textbf{48.39}\\ 36.01\\ 38.70 \\40.85 } & \makecell[c]{\textbf{53.78}\\ 37.51\\ 36.61 \\52.13 } & \makecell[c]{31.25\\ 18.18\\ 20.00 \\\textbf{80.00} } & \makecell[c]{23.81\\ 9.52\\ 19.04 \\ \textbf{64.51} } & \makecell[c]{27.03\\ 12.5\\ 19.51 \\ \textbf{71.42} }\\
        \bottomrule
    \end{tabular}
    \caption{Model performances for second-level (L2) and third-level (L3) multi-label classifications.}
    \label{tab:result-multi}
\end{table*}

Along with the datasets, we provide baseline quantitative results obtained by the chosen models.

The model performances for the single-label classification datasets are given in \Cref{tab:result-single}. For the L1 datasets, all models perform similarly for both Ottoman and Russian. On the other hand, on the L2 datasets, BoW + NB performs significantly better than other models on Ottoman, and also compared to BoW + NB on Russian L2. This might be due to the fact that literary text types of articles especially for Ottoman can be easily categorized by the certain words they have. However, there is no particular model that outperforms others in all cases. These single-label classification tasks are on higher-level datasets, namely L1 and L2, and are easier tasks compared to the lower-level and multi-label counterparts. Although the model specifications greatly vary (e.g. a billion-parameter model vs. a classical BoW + NB model), the results are comparable among the models for the single-label classification tasks. 

The results obtained from the multi-label datasets are given in \Cref{tab:result-multi}. Similar to single-label classification results, simple BoW + NB model performs sufficiently well and surpasses its LLM counterparts on Ottoman L2 and L3 datasets. For Russian datasets, Llama consistently performs as the best model in almost all tasks with the only exception being Russian L3 literature-subject-genre where BoW + NB outperformed Llama-2. This is probably because of the pre-training data containing a portion of Russian content. Also for mBERT, we see that the metrics at 0.5 threshold namely, AP, AR, and AF1 are 00.00\% for several datasets signaling that the model's confidence scores for multi-label classification tasks do not even attain the 50\% cut-off. 

Llama-2 and Falcon are mostly on par for Ottoman Turkish datasets, yet for Russian datasets Llama-2 outperforms Falcon in most cases. Several factors affecting this could be that Llama-2-7b and Falcon-7b were pre-trained on a corpus of 2T with 32K vocabulary size and 1.5T tokens with 65K vocabulary size, respectively.

\section{Discussion}
\label{sec:discussion}


The experiments show that a simple model like BoW+NB can be comparable to prominent models like Llama-2 and Falcon in a low-resource setting. However, it is important to point out that Llama-2 and Falcon, which have billions of parameters, underwent training with a frozen backbone and 4-bit integer quantization. This approach can substantially hurt the performance of the models.

Based on mAP (weighted), Llama-2 is the best-performing model in both languages in level-1 (L1) tasks, where our dataset has the largest sample size and is well-balanced. For level-2 (L2) and level-3 (L3) tasks, BoW+NB outperforms Llama-2, Falcon, and mBERT in Ottoman classification tasks; yet in Russian, the leading model is still Llama-2. In Russian, the leading model remains Llama-2 in two of the three classification tasks of L3, while BoW+NB is the best-performing model in one of the tasks. Here, it is important to note that, as the sample size gets smaller, we observe the performance difference between BoW+NB and Llama-2 decreases. The results show that, in an imbalanced setting, different metrics elect different models. Due to the nature of our dataset, we can observe this more clearly at lower-level (L2, L3) classification tasks. 

At lower-level categories, the classification tasks become more challenging, both in terms of generating a sufficient number of instances and establishing coherent and objective labeling across experts. For example, the subcategories of the Literature/ Movements labels are open to interpretation, over which experts might disagree. These subcategories, which signify the subject of any critical article on literature, are fluid categories that may overlap in many instances.

In our dataset annotation process detailed in \Cref{ssec:labeling}, we opted against the traditional cross-check annotation, and did not calculate inter-annotator agreement (IAA) scores due to the specific nature of our annotation task. Our annotation teams consist of experts proficient in the relevant languages and the historical context of the 18\textsuperscript{th} and 19\textsuperscript{th} centuries. Unlike tasks such as question answering or sentiment analysis, our categorization process is not inherently subjective but requires specialized expertise. Also, comprehensive guidelines and training were provided to annotators, and regular consensus meetings were held to resolve challenging cases. Consequently, we implemented a hierarchical cross-check system: initial annotations by a team member were reviewed by a peer expert, and then revised by the team leader, whose domain knowledge surpassed that of the other annotators, overwriting the previous annotations. This structure ensured high-quality annotations and rendered traditional IAA metrics less meaningful.

\section{Conclusion}
\label{sec:conclusion}

This paper proposes a collection of carefully curated datasets of some prominent historical non-Western languages, along with a set of benchmarks obtained from contemporary state-of-the-art LLMs and the BoW naive Bayes model. We observed the shortcomings of current leading LLMs in handling downstream tasks with low-resource languages that are not included in their pre-training data. Notably, we also observed a minimal performance difference between BoW and LLMs. This indicates a need for further research on LLMs. We believe that the democratization of contemporary NLP efforts requires the inclusion of low-resource languages. Even if the majority of the population does not use these languages, this line of research can provide significant assistance to historians, linguists, and humanities scholars. 

\section*{Acknowledgements}
This article received funding from the European Research Council (ERC) project NONWESTLIT under the European Union’s Horizon 2020 research and innovation program (Grant Agreement No. 950513).  Dr. Akbas is supported by the ``Young Scientist Awards Program (BAGEP)" of Science Academy, Türkiye. We would like to express our sincere gratitude to the following individuals for their invaluable contributions to this project: Mehtap Öztürk, Zeynep Nur Şimşek, Hazal Bozyer, Jennifer Flaherty, Nikita Drozdov, Ilya Kliger and Kiril Zubkov. 


\section*{Limitations}

There are three major challenges to this work. First, the data collection process requires experts and significant human effort. All articles must be labeled by human experts. They are required to read each article in detail and categorize it accordingly; however, the variance in the experts' judgments needs to be handled by cross-checking. 
In addition, there are currently no digital resources or OCR programs for the Ottoman Turkish language that can perform at an acceptable level without excessive human effort.

Second, in terms of NLP research, the dataset itself poses several challenges. The limited amount of data hinders the fine-tuning process. The articles are very long, requiring their division into chunks. Classification is performed by averaging the predicted probabilities of each 2048-token segment, but most of the segments may not have enough content to classify the whole article into the correct category. 
 
The third limitation lies in the cultural field of the period. Although experts pay considerable attention to capturing a wide representation of formal and thematic properties of the period, which is represented in the taxonomy, it is impossible to have an equal or fair distribution among labels. The cultural and aesthetic world has its sui generis complexities that resist taxonomies and quantitative analysis.

\bibliography{custom}

\appendix

\section{Dataset Structure}
\label{apx:structure}

\begin{figure*}[ht]
\centering
\includegraphics[width=15cm]{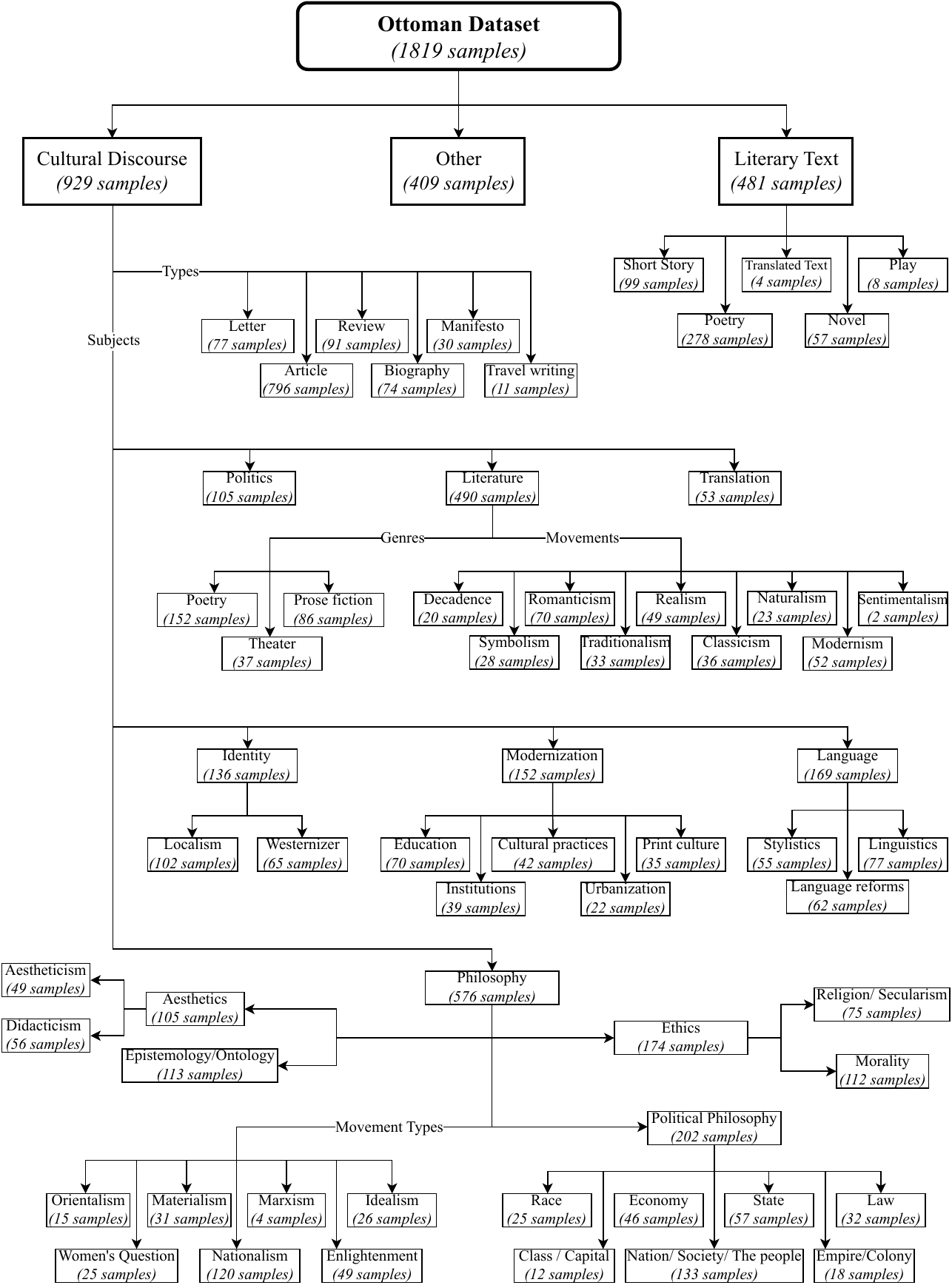}
\caption{Number of samples in each category for Ottoman Dataset}
\centering
\label{fig:ottoman_dat}
\end{figure*}

\begin{figure*}[ht]
\centering
\includegraphics[width=15cm]{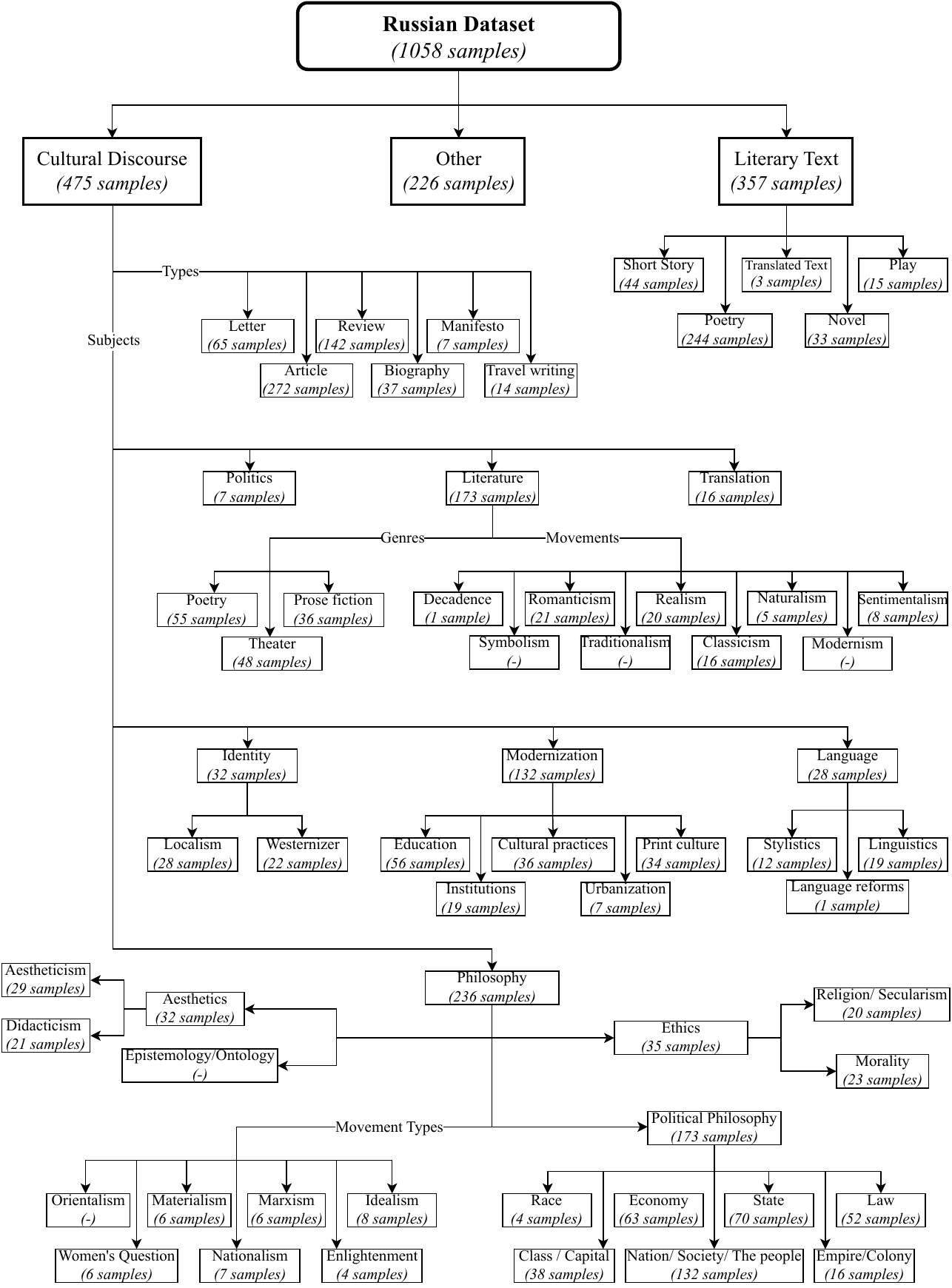}
\caption{Number of samples in each category for Russian Dataset}
\centering
\label{fig:russian_dat}
\end{figure*}

The dataset structure displaying the hierarchy is given in \Cref{fig:ottoman_dat} for the Ottoman Turkish dataset and \Cref{fig:russian_dat} for the Russian dataset.

\end{document}